\documentclass{llncs}

\usepackage{verbatim,color,latexsym,amssymb,amsmath,algorithm2e}%,multirow}

\newcommand{\DNote}[1]{\ifthenelse{\boolean{draftmode}}{\medskip{\noindent{\bf{Note}:} {\em #1}}\medskip}{}}
\definecolor{darkblue}{rgb}{0.1,0.1,0.5}
\definecolor{vdarkblue}{rgb}{0.05,0.05,0.4}
\definecolor{midgrey}{rgb}{0.5,0.5,0.5}
\definecolor{darkgrey}{rgb}{0.1,0.1,0.1}
\definecolor{darkred}{rgb}{0.7,0.1,0.1}
\definecolor{midred}{rgb}{0.5,0.1,0.1}
\definecolor{midgreen}{rgb}{0.3,0.5,0.3}
\definecolor{darkgreen}{rgb}{0.1,0.3,0.1}

\newcounter{marginote}
\newcommand{\todoF}[2]{}

\begin{document}

\title{On Validating Boolean Optimizers}

\author{Antonio Morgado\inst{1} \and Joao Marques-Silva\inst{1,2}}

\institute{CSI/CASL,UCD, Dublin, Ireland
\and IST/INESC-ID, TUL, Lisbon, Portugal\\
\email{\{ajrm,jpms\}@ucd.ie}}

\maketitle

%------------------------------------------------------------------------------%
% File:        abs.tex
%------------------------------------------------------------------------------

\begin{abstract}
Boolean optimization finds a wide range of application domains, that
motivated a number of different organizations of Boolean optimizers
since the mid 90s.
%
%Among these,
%Moreover, 
Some of the most successful approaches are based on iterative calls to
an NP oracle, using either linear search, binary 
search or the identification of unsatisfiable sub-formulas.
The increasing use of Boolean optimizers in practical settings raises
the question of confidence in computed results. For example, the issue
of confidence is paramount in safety critical settings.
One way of increasing the confidence of the results computed by
Boolean optimizers is to develop techniques for validating the
results.
Recent work studied the validation of Boolean optimizers
% in the context of 
based on 
branch-and-bound search~\cite{larrosa-jar09,larrosa-sat09}.
This paper complements existing work, and develops methods for
validating Boolean optimizers that are based on iterative calls to an
NP oracle.
This entails implementing solutions for validating both satisfiable
and unsatisfiable answers from the NP oracle. The work described in
this paper can be applied to a wide range of Boolean optimizers,
that find application in Pseudo-Boolean Optimization and in Maximum
Satisfiability.
Preliminary experimental results indicate that the impact of the
proposed method in overall performance is negligible.
%
%
\begin{comment}
ajrm:: throwing ideas

Boolean optimization solvers have been used with success in many
applications.
%
In recent years, different types of Boolean optimization solvers have appeared.
%
Some of the most succesfull approaches include iteration of SAT
instances and are based in linear search, binary search or in the
identification of unsatisfiable cores.
%
Having confidence on the solution reported by the
optimization solver is paramount for some applications.
%
Thus the need for a way to validate the results reported by the
Boolean optimization solver.
%
Previously validation of Boolean optimization solvers has been done in
the context of branch-and-bound\cite{larrosa-jar09,larrosa-sat09}.
%
In this paper we propose a method to validate Boolean optimization
solvers that are based on iterative approaches.
%
The method is based on validating iterations of the solver with
satisfiable answers as well as unsatisfiable ones.
%
The proposed method is applied to one Boolean Optimization problem,
the Max SAT problem, and was implemented in msuncore. 
\end{comment}
% 
\end{abstract}

%------------------------------------------------------------------------------%

%------------------------------------------------------------------------------%
% File:        intro.tex
%
% Description: Introduction.
%
% Author:      Joao Marques-Silva.
%
% Created:     07 May 2009.
%------------------------------------------------------------------------------%

\section{Introduction}
\label{sec:intro}

% to remove just bullets with overall ideas of paragraphs----------------------%
%\hrule
%{\scriptsize PARAGRAPHS AS BULLETS \input{bullets.intro}}
%\hrule
%------------------------------------------------------------------------------%

The remarkable advances of Boolean Satisfiability (SAT) algorithms in
past years, motivated their widespread use in many practical
applications. 
Moreover, some applications require the Boolean algorithms to optimize
some 
%value or a 
cost function (e.g. \cite{jose-pldi11}, \cite{safarpour-fmcad07}, \cite{roussel-hbs09}, \cite{li-hbs09}).
%
%For example, a partial MaxSAT solver has been used for locating a
%potential bug in the context of design
%debugging~\cite{jpms-glsvlsi09}.

These practical applications depend on correct results computed by
Boolean optimizers to fulfill their objective.
One way of increasing the confidence in the results computed, is to
develop techniques for validating the results.
In the context of (pure) SAT, validation of unsatisfiable answers has
been addressed by Zhang \& Malik~\cite{zhang-date03}, whereas the
validation of satisfiable answers corresponds just to a check if the
returned assignment satisfies all the clauses.
Certified validation of SAT has been considered in ~\cite{lescuyer-tphol08}.

Validation of Boolean optimizers has been studied
recently~\cite{larrosa-jar09,larrosa-sat09}. However, this work covers 
only Boolean optimizers that are based on branch-and-bound search,
e.g.~BSOLO~\cite{jpms-amai04}.
Nevertheless, many state of the art Boolean optimizers are based on
iterative calls to a SAT solver.
For example, minisat+~\cite{een-jsat06}, sat4j~\cite{sat4j}
(-pb,-maxsat), PBS~\cite{ARMS02}, pueblo~\cite{kas-jsat06},
msuncore~\cite{jpms-date08,jpms-sat08b,jpms-sat09},
PM2~\cite{bonet-sat09a}, WPM1~\cite{bonet-sat09a},WMSU1~\cite{jpms-sat09}.

This paper develops methods for validating the results computed by
Boolean optimizers, that are based on iterative calls to a SAT
solver.
The idea is to allow for an independent checker to receive the
information returned by the Boolean optimizers and validate the
result.
The paper shows that, to improve the efficiency of the solution
checking process, it is unnecessary for the checker to validate all
the iterations of the solver, which can be as many as the number of
clauses. This result is general and, as the paper shows, holds for
most Boolean optimization algorithms based on iterative calls to a SAT
solver.
Moreover, the paper shows that, similarly to SAT, the time to check
MaxSAT solutions is in general negligible when compared with the time 
the algorithms take to compute the optimum solution.

The paper is organized as follows.
Section~\ref{sec:prelim} introduces Boolean optimization problems
and describes three iterative approaches currently used by Boolean
optimizers. 
Section~\ref{sec:validating} presents our methods of validation
of the results returned.
Experimental results with our methods of validation are shown in
Section~\ref{sec:res}, and the paper concludes in Section~\ref{sec:conc}.

% Section~\ref{sec:prelim} introduces Boolean optimization problems
% and shows some transformations between instances of the problems.
% %
% Next, Section~\ref{sec:algorithms} describes three iterative
% approaches currently used by Boolean optimization solvers, and
% establishes the main results used for validating the results computed
% by Boolean optimization solvers.
% %
% Section~\ref{sec:validating} presents our method of validation
% of the results returned.
% %
% Experimental results with our method of validation are shown in
% Section~\ref{sec:res}, and the paper concludes in Section~\ref{sec:conc}.

%------------------------------------------------------------------------------%

%------------------------------------------------------------------------------%
% File:        prelim.tex
%
% Description: Preliminaries.
%
% Author:      Joao Marques-Silva.
%
% Created:     07 May 2009.
%------------------------------------------------------------------------------%
\section{Preliminaries}
\label{sec:prelim}

% to remove just bullets with overall ideas of paragraphs----------------------%
%\hrule
%{\scriptsize PARAGRAPHS AS BULLETS \input{bullets.prelim}}
%\hrule
%THE ABOVE IS TO REMOVE\\
%------------------------------------------------------------------------------%

This section describes Boolean optimization problems and the
notation used in the paper.
First, Boolean optimization problems are introduced, followed by a
brief introduction to the different iterative algorithms to solve
Boolean optimization problems.
Familiarity with the basic concepts of Boolean variables, literals,
clauses, conjunctive normal formula (CNF), etc, is assumed, but
additional detail can be found for example in~\cite{SatHandbook09}.
A CNF formula can be viewed either as a conjunction of clauses or as a 
set of clauses. Similarly, a clause can be viewed as a disjunction of
literals or as a set of literals.

The Boolean optimization problems described in this section can be
seen as optimization extensions of the propositional satisfiability
problem (SAT). 
The SAT problem is the problem that, given a CNF formula $\varphi$ on a set of
Boolean variables, determines one assignment $\sigma$ to the variables
such that $\sigma$ satisfies $\varphi$, if such $\sigma$ exists.
Otherwise \emph{UNSATISFIABLE} is reported.

This work considers three different Boolean optimization
problems. 
The MinCost SAT problem~\cite{fu-iccad06}, the Pseudo-Boolean
optimization problem (PBO) \cite{roussel-hbs09} and the
Maximum Satisfiability problem (MaxSAT) 
\cite{li-hbs09}. 
The MinCost SAT problem has as input a CNF formula 
$\varphi\equiv\bigwedge_{j=1}^m\omega_j$, and a cost function represented as in
Equation~(\ref{eq:prelim:mincostsat-def}), where each $c_i$ is a constant
and literal $l_i$ is either variable $x_i$ or the negated variable
$\overline{x_i}$. 
The goal in MinCost SAT is to determine a satisfying assignment $\sigma$ 
for $\varphi$ that minimizes the value of the cost function. 
\begin{equation}
  \begin{array}{rl}
     \min: \sum_{i=1}^nc_ia_i\textrm{ where }
     \left\{\begin{array}{l}a_j=1\textrm{ if
         }\sigma(l_i)=true\\a_j=0\textrm{
           otherwise}\end{array}\right.\\
%
%    \textrm{Cost Function} & \sum_{i=1}^nc_il_i\\
%    \\
%    \textrm{Set of Clauses} & \{\omega_j ~|~ 1\leq j\leq m\}
  \end{array}
  \label{eq:prelim:mincostsat-def}
\end{equation}

The Pseudo-Boolean Optimization problem (PBO) also considers a cost
function  (also called the objective function), but PBO does
not consider the formula $\varphi$ to be in CNF format.
Instead in PBO, the constraints are linear\footnote{Some of the current
  work in PBO does not require the constraints or the objective
  function to be linear. 
  The linear formulation is preferred for readability, but
  non-linear constraints or non-linear objective function can be considered.} 
inequations of pseudo-Boolean variables, that is, the variables have
either value $0$ or $1$.
An instance of PBO can be formulated as in
Equation~(\ref{eq:prelim:pbo-def}). 
\begin{equation}
  \begin{array}{rlp{1cm}l}
    \min: & \sum_{i=1}^nc_ix_i&&1\leq j\leq m\\
    &&&a_{i,j},b_j,c_i\in \mathbb{Z}\\
    \textrm{s.t.} & \sum_{i=1}^na_{i,j}l_{i,j} \leq b_j&&l_i,l_{i,j}\in \{x_i,\overline{x_i}\}\\
    &&&x_i \in\{1,0\}

  \end{array}
  \label{eq:prelim:pbo-def}
\end{equation} 

%Given a CNF formula $\varphi$, the MaxSAT problem consists in
%identifying the assignment $\sigma$ to the Boolean variables of
%$\varphi$ that satisfies the maximum number of clauses of $\varphi$.
%
The last Boolean Optimization problem considered is the Maximum
Satisfiability (MaxSAT) problem
which given a CNF formula $\varphi\equiv\bigwedge_{j=1}^m\omega_j$
consists on identifying an assignment $\sigma$ that satisfies the
maximum number of clauses of $\varphi$. 
MaxSAT can be represented as in Equation~\ref{eq:prelim:maxsat-def}.
Typically in MaxSAT the given formula $\varphi$ is unsatisfiable (otherwise
MaxSAT corresponds to the SAT problem).
Note also that despite the MaxSAT problem being defined in terms of
the maximum number of satisfied clauses, current MaxSAT solver report
as solution the minimum number of unsatisfied clauses\footnote{In the
  pseudo-codes presented it is the minimum number of unsatisfied
  clauses that is returned.}.
The maximum number of satisfied clauses is then obtained by subtracting
from the total number of clauses, the minimum number of unsatisfied clauses.
\begin{equation}
  % \begin{array}{rl}
  \max:  \sum_{j=1}^ma_j,\quad  \left\{\begin{array}{l}a_j=1\textrm{
        if }\sigma(\omega_j)=true\\a_j=0\textrm{
        otherwise}\end{array}\right.
  % \\
  % \textrm{s.t.} & \varphi\equiv\bigwedge_{j=1}^m\omega_j
  % \end{array}
  % 
  % \max_{\sigma} \sum \{a_j|a_j=1\textrm{ if }\sigma(\omega_j)=true\textrm{,
  %   $a_j=0$ otherwise}\} 
  \label{eq:prelim:maxsat-def}
\end{equation}

Several variants of MaxSAT can be considered, namely \emph{partial} MaxSAT,
\emph{weighted} MaxSAT and \emph{weighted partial} MaxSAT.
In partial MaxSAT some of the clauses on $\varphi$, called the
\emph{hard} clauses, must be satisfied by $\sigma$.
By opposition, the clauses of $\varphi$ which may or may not be
satisfied by $\sigma$ are called \emph{soft} clauses.
Thus, the objective in partial MaxSAT is to determine $\sigma$ that
satisfies all the hard clauses of $\varphi$, and maximizes the number
of satisfied soft clauses of $\varphi$, if such $\sigma$ exists.
If no $\sigma$ is able to satisfy the hard clauses of $\varphi$, then
\emph{UNSATISFIABLE} is returned. 

In weighted MaxSAT each clause is associated to a weight that represents
the cost of satisfying the clause.
The objective is to maximize the sum of the weights of the satisfied clauses.
Weighted partial MaxSAT combines the previous two.
The set of clauses is divided in hard and soft clauses, and the soft
clauses are associated with weights.
The objective is to maximize the sum of the weights of the satisfied
soft clauses while satisfying all the hard clauses.

This paper focus on (pure) MaxSAT and partial MaxSAT.

Although the Boolean optimization problems have different
optimization objectives and different formalisms, it is possible to translate an
instance of one problem into the
others~\cite{dlb-ictai96,kas-iccad02,HLO08,jpms-sat09}.

Despite only considering these Boolean optimization problems, the
methods proposed in this paper for checking computed results can be
adapted to other Boolean optimization problems, such as
MaxSMT~\cite{nieuwenhuis-sat06,cimatti-tacas10},
MaxASP~\cite{emilia-lpnmr09} or Weighted Boolean Optimization
(WBO)~\cite{jpms-sat09}.

%------------------------------------------------------------------------------%
%------------------------------------------------------------------------------%
% File:        algorithms.tex
%
% Description: Boolean Optimization Algorithms and Approaches.
%
% Author:      Antonio Morgado.
%
% Created:     27 April 2010.
%------------------------------------------------------------------------------%

\subsection{Boolean Optimization Algorithms}
\label{sec:algorithms}

In this section we outline three approaches to solve Boolean
optimization problems for which in the next section we apply our method of
validation.
The approaches are based on linear search, binary search and
(unsatisfiable) core guided search.
Given that translations between the Boolean optimization problems are
known, in this section, we concentrate on the problem of MaxSAT for
presenting the different algorithms.

The idea of the algorithms presented is to start with
a bound for the optimum value and create a CNF instance that
improves over the given bound.
Depending on the satisfiability of the created CNF instance, the bound
is updated and the process is restarted until the optimum value is found.

Considering the case of MaxSAT, the bound corresponds to the number of clauses
that are unsatisfiable.
The optimum value is then obtained by subtracting the bound from the
total number of clauses.
The algorithms consider integer variables $\mu$/$\lambda$ to hold the
 value of the upper/lower bound (respectively) depending on the type of
search.

All the three approaches consider \emph{relaxation variables} which are new
fresh variables that are added to the clauses of the MaxSAT instance.
If $r_i$ is the relaxation variable associated with clause $\omega_i$,
then $r_i$ is assigned to \emph{true} if $\omega_i$ is unsatisfiable.
Otherwise, $r_i$ is assigned to \emph{false}.

%------------------------------------------------------------------------------%

\subsubsection{Linear Search}
\label{sec:alg-linear-search}

\IncMargin{1em} 
\begin{algorithm}[t]
  \SetKwData{True}{\emph{true}}\SetKwData{sat}{SATISFIABLE} 
  \SetKwFunction{Sat}{SAT}\SetKwFunction{Cnf}{CNF}
  \SetKwInOut{Input}{input}\SetKwInOut{Output}{output} 
  \Input{A CNF formula $\varphi = \omega_1 \wedge \ldots \wedge
    \omega_n$}  
  %\Output{Maximum number of simultaneously satisfiable clauses of $\varphi$}
  \BlankLine 
  $R \leftarrow \{r_i:r_i$ associated with $\omega_i\in
  \varphi\}$\;
  $\varphi_W  \leftarrow \bigwedge_{i=1}^{n}(\omega_i\vee
  r_i)$\;
  $\lambda \leftarrow 0$\;
  \While{$\True$}{ 
    $\varphi^{card} \leftarrow$ \Cnf{$\sum_{r_i\in R}r_i \leq \lambda$}\; 
    $st\leftarrow $ \Sat{$\varphi_W~ \cup~ \varphi^{card}$}\;
    \eIf{$st=$ \sat}{
      \Return{$\lambda$}\;
    }{
      $\lambda \leftarrow \lambda+1$\;
    } 
  }
  \caption{Linear Search unsatisfiable-satisfiable for MaxSAT}\label{alg:lsus} 
\end{algorithm}\DecMargin{1em}

\IncMargin{1em} 
\begin{algorithm}[t]
  \SetKwData{True}{\emph{true}}\SetKwData{sat}{SATISFIABLE} 
  \SetKwFunction{Sat}{SAT}\SetKwFunction{Cnf}{CNF}
  \SetKwInOut{Input}{input}\SetKwInOut{Output}{output} 
  \Input{A CNF formula $\varphi = \omega_1 \wedge \ldots \wedge
    \omega_n$}  
  %\Output{Maximum number of simultaneously satisfiable clauses of $\varphi$}
  \BlankLine 
  $R \leftarrow \{r_i:r_i$ associated with $\omega_i\in
  \varphi\}$\;
  $\varphi_W \leftarrow \bigwedge_{i=1}^{n}(\omega_i\vee
  r_i)$\; 
  $\mu \leftarrow n$\;
  \While{$\True$}{ 
    $\varphi^{card} \leftarrow$ \Cnf{$\sum_{r_i\in R}r_i \leq \mu$}\; 
    $(st,\sigma)\leftarrow $\Sat{$\varphi_W~ \cup~ \varphi^{card}$}\;
    \eIf{$st=$ \sat}{
      $\mu \leftarrow \sum_{i=1}^{n}\sigma(r_i)$\;
    }{
      \Return{$\mu$}\;
    } 
  }
  \caption{Linear Search satisfiable-unsatisfiable for MaxSAT}\label{alg:lssu} 
\end{algorithm}\DecMargin{1em}

The algorithms described in this section perform a linear
search on the optimum value.
The use of Linear search for MaxSAT was first proposed in 2006~\cite{FM06}.

A new working CNF formula $\varphi_W$ is created to
contain all the clauses of the form $(\omega_i\vee r_i)$, where
$\omega_i$ is a clause of the MaxSAT instance and $r_i$ is the
relaxation variable associated to $\omega_i$.
Each iteration adds a cardinality constraint to $\varphi_W$ to
constrain the number of relaxation variables assigned to
\emph{true} to be at most a given value.
The exact value used in the cardinality is dependent on the type of search.
Two types of linear search are possible.
Either searching through unsatisfiable CNF instances or searching
through satisfiable CNF instances.
The case of searching through unsatisfiable CNF instances is called
\emph{Linear Search unsatisfiable-satisfiable}, and
the algorithm starts by seting a lower bound $\lambda$ to 0.
Each iteration for which an unsatisfiable instance is found,
$\lambda$ is increased by one. 
The search proceeds until a satisfied instance is found.
The pseudo-code is presented in Algorithm~\ref{alg:lsus}.

The case of searching through satisfiable CNF instances is called
\emph{Linear Search satisfiable-unsatisfiable}. 
The algorithm uses an upper bound $\mu$ which initially is set 
to the total number of clauses.
Each iteration, for which $\varphi_W$ is found satisfiable, decreases
$\mu$ by the number of relaxation variables assigned to true.
The number of relaxation variables assigned to true is obtained through
the assignment $\sigma$ returned by the SAT solver on satisfiable instances. 

The search stops when an unsatisfiable instance is found.
The pseudo-code is presented in Algorithm~\ref{alg:lssu}.
An example of a MaxSAT solver that uses Linear Search satisfiable-unsatisfiable is SAT4j-maxsat~\cite{sat4j}.

%------------------------------------------------------------------------------%

\subsubsection{Binary Search}
\label{sec:alg-bin-search}

\IncMargin{1em} 
\begin{algorithm}[t]
  \SetKwData{True}{\emph{true}}\SetKwData{sat}{SATISFIABLE} 
  \SetKwFunction{Sat}{SAT}\SetKwFunction{Cnf}{CNF}
  \SetKwInOut{Input}{input}\SetKwInOut{Output}{output} 
  \Input{A CNF formula $\varphi = \omega_1 \wedge \ldots \wedge
    \omega_n$}  
  %\Output{Maximum number of simultaneously satisfiable clauses of $\varphi$}
  \BlankLine 
  $R \leftarrow \{r_i:r_i$ associated with $\omega_i\in \varphi\}$\;
  $\varphi_W \leftarrow \bigwedge_{i=1}^{n}(\omega_i\vee r_i)$\;
  $\mu \leftarrow n$\;
  $\lambda \leftarrow -1$\;
  \While{$(\mu >\lambda_{bot}+1)$}{ 
    $\tau \leftarrow \lfloor\frac{\mu+\lambda}{2}\rfloor$\; 
    $\varphi^{card} \leftarrow $ \Cnf{$\sum_{r_i\in R}r_i \leq \tau$}\;
    $(st,\sigma)\leftarrow $ \Sat{$\varphi_W~ \cup~ \varphi^{card}$}\; 
    \eIf{$st=$ \sat}{
      $\mu \leftarrow \sum_{i=1}^{n}\sigma(r_i)$\;
    }{
      $\lambda \leftarrow \tau$\;
    } 
  }
  \Return{$\mu$}\;
  \caption{Binary search for MaxSAT}\label{alg:binary} 
\end{algorithm}\DecMargin{1em}

The algorithm presented in this section is similar to the Linear
search algorithms, but making instead a binary search on the value of
the bound.

%combines the two linear
%search algorithms of Section~\ref{sec:alg-linear-search}.
%
Binary Search uses both an upper bound $\mu$ and and a lower bound
$\lambda$, and iteratively creates and solves a CNF instance that
includes a constraint relating the cost function with a middle value
$\tau$.

%Consider again the case of MaxSAT.
%
%The lower and the upper bounds are the same as the bounds used in the
%previous algorithms.

The algorithm is shown in Algorithm~\ref{alg:binary}.
The relaxation variables and the working formula are created as
in the previous linear algorithms.
%
%Two $\lambda$'s are needed, $\lambda_{bot}$ for an upper
%bound, and $\lambda_{top}$ for a lower bound
%
%\footnote{The upper and
%  lower bound refer to the number of clauses satisfied, whereas the
%  $\lambda$'s refer to the number of relaxation variables assigned
%  \emph{true}.}.
%
In each iteration the middle value of
$[\lambda,\mu]$ is assigned to $\tau$ and
the set of clauses that constrains the sum of the relaxation variables
to be at most $\tau$ is feed to the SAT solver, together with the
working formula. 
If the SAT solver returns {\em SATISFIABLE}, then $\mu$ is
updated to the number of relaxation variables assigned to true by the
assignment $\sigma$ returned by the SAT solver.
Otherwise, $\lambda$ is updated to $\tau$.
The search stops when $\mu$ and $\lambda$ differ in
one unit, in which case, the maximum number of satisfiable clauses
is the number of original MaxSAT clauses minus $\mu$.

%As was suggested for linear search over satisfiable instances, an
%improvement can be when the SAT solver returns {\em SATISFIABLE}.
%
%Instead of updating $\lambda_{top}$ to the value of $\lambda_{mdl}$,
%$\lambda_{top}$ can be updated to the total number of relaxation
%variables assigned \emph{true} by the satisfiable assignment returned
%by the SAT solver.

%The use of binary search for Boolean optimization is an immediate
%extension of the basic linear search algorithms. For example, 
The use of binary search for PBO has been
discussed~\cite{SatHandbook09}. Moreover, binary search has been
recently used for solving Boolean optimization 
problems in the context of SMT~\cite{cimatti-tacas10}, where the
authors developed a theory of costs SMT($\mathcal{C}$) and
SMT($\mathcal{C}\cup\mathcal{T}$), and propose to solve PBO, MaxSAT
and MaxSMT by encoding the problems into SMT($\mathcal{C}$) (and
SMT($\mathcal{C}\cup\mathcal{T}$) for MaxSMT).

%------------------------------------------------------------------------------%

\subsubsection{Core Guided Search}
\label{sec:alg-unsat-searxh}

%The algorithms of Boolean optimization that are based on
%unsatisfiability search rely on the concept of unsatisfiable cores.
%
Another type of search used for Boolean optimization is based on the
generation of unsatisfiable cores.
An \emph{unsatisfiable core} (or simply \emph{core}) is a sub-formula
of the original CNF formula that is unsatisfiable~\cite{zhang-date03}.
Current SAT solvers are able to return cores for
unsatisfiable instances (which are regarded as a reason for the
unsatisfiability of the instance).

\IncMargin{1em} 
\begin{algorithm}[t]
  \SetKwData{True}{\emph{true}}\SetKwData{sat}{SATISFIABLE} 
  \SetKwFunction{Sat}{SAT}\SetKwFunction{Cnf}{CNF}
  \SetKwInOut{Input}{input}\SetKwInOut{Output}{output} 
  \Input{A CNF formula $\varphi = \omega_1 \wedge \ldots \wedge \omega_n$} 
  %\Output{Value $z^*$ maximum number of clauses of $\varphi$ simultaneously satisfiable }  
  %\Output{Minimum number of unsatisfiable clauses of $\varphi$}
  \BlankLine 
  $R \leftarrow \emptyset$\;
  $\varphi_W \leftarrow \varphi$\;
  $\lambda \leftarrow 0$\;
  \While{\True}{
    $\varphi^{card} \leftarrow $ \Cnf{$\sum_{r\in R}r\leq\lambda$} \;
    $(st,\varphi_C)\leftarrow $\Sat{$\varphi_W~ \cup~ \varphi^{card}$}\;
    \eIf{$st=$ \sat}{
      \Return{$\lambda$}\;
    }{
      $\lambda \leftarrow \lambda+1$\;
      \ForEach{$\omega\in\varphi_C$}{
        \If{$\omega$ has no relaxation variable}{
          $r$ is a new relaxation variable\\
          $R\leftarrow R\cup\{r\}$\;
          $\omega_r\leftarrow \omega \cup \{r\}$\;
          $\varphi_W\leftarrow\varphi_W\setminus\{\omega\}\cup\{\omega_r\}$\;
        }
      }
    } 
  }
  \caption{Simplified MSU3 }\label{alg:msu3} 
\end{algorithm}\DecMargin{1em} 

The use of unsatisfiable cores for solving (partial) MaxSAT was first
proposed in 2006~\cite{FM06}
%, in what will be referred to as the
with the MSU1.0 algorithm.
The idea of MSU1.0 is to iteratively eliminate unsatisfiable cores of the
problem instance, computed by a SAT solver, by adding new relaxation
variables to the clauses and add new constraints to constrain the sum
of these relaxation variable to be equal to one.

%MSU1.0 iteratively eliminates unsatisfiable cores of the
%problem instance, computed by a SAT solver.
%
%In each iteration, if the SAT solver reports {\em UNSATISFIABLE} together
%with a new unsatisfiable core, then each clause $w_i$ of the core
%(corresponding to an original clause)
%is replaced by the same clause
%with the addition of a fresh relaxation variable $r_i$.
%
%Then the constraint $CNF(\sum_{r_i\in RV(core)}r_i=1)$
%that encodes the sum of all relaxation variables
%(added by the current core) to be exactly one 
%is added to the set of all clauses, where $RV(core)$ is the set of
%relaxation variables associated to the clauses of the core.
%
%This constraint is referred as an \emph{One-Hot} constraint. It is
%composed of two distinct constraint, one denoting the AtLeast1
%constraint, and the other denoting the AtMost1 constraint. The
%encoding of the AtMost1 constraint used is commonly known as the
%{\em pairwise encoding}~(e.g.~\cite{gent-ecai02}).

Recently, several new
MSU algorithms~\cite{jpms-sat08b,jpms-date08},
PM2 algorithm~\cite{bonet-sat09a}, WPM2~\cite{ansotegui-aaai10},
bin-core and bin-core-disjoint~\cite{heras-aaai11} have been proposed. 
The differences of the algorithms include the number of cardinality
constraints used, the encoding of the cardinality constraints, the
number of relaxation variables considered for each clause and how the
algorithm proceeds (despite being all based on the generation of cores).

In the following we consider a representative algorithm for this class
of algorithms.
We consider a simplified version of the
MSU3~\cite{jpms-date08} algorithm 
which is depicted in
Algorithm~\ref{alg:msu3}.
Instead of relaxing all the (soft) clauses as in previous algorithms,
the set of clauses that are allowed to be relaxed in one iteration is
dependent on the cores reported in the previous iterations.
Initially, no clause is allowed to be relaxed, thus the set of all relaxed
variables $R$ is empty and the working formula $\varphi_W$ is the same as the
original formula.
%
%The algorithm also assumes, as in linear search (through unsatisfiable
%instances), that in the beginning no clause is unsatisfiable, thus
%$\lambda$ is set to $0$. 

In each iteration the SAT solver is called with the working formula
and an AtMost constraint on the number of relaxed variables.
If the SAT solver returns {\em UNSATISFIABLE} then a new core
$\varphi_C$ is available.
$\lambda$ is increased by one and each original (soft) clause in the
core receives a new relaxation variable, if it does not have one yet.
Otherwise, the optimum has been found.

%One final remark is that the linear search algorithm, where search is
%made through unsatisfiable instances, can be viewed as a particular 
%case of this simplified MSU3 algorithm, but where the unsatisfiable
%core returned by the SAT solver is always the full formula; or the
%other way around. 
%
%The simplified MSU3 algorithm corresponds to a linear search algorithm,
%with the enhancement of core guidance on which clauses to relax. 

%------------------------------------------------------------------------------%

%------------------------------------------------------------------------------%
% File:        validating.tex
%
% Description: Validating Boolean Optimization Solvers.
%
% Author:      Antonio Morgado.
%
% Created:     27 April 2010.
%------------------------------------------------------------------------------%

%\section{Validating Boolean Optimization Solvers}
\section{Validating Boolean Optimizers}
\label{sec:validating}

% to remove just bullets with overall ideas of paragraphs----------------------%
%\hrule
%{\scriptsize PARAGRAPHS AS BULLETS \input{bullets.validating}}
%\hrule
%------------------------------------------------------------------------------%

Validating the results provided by a solver is a recurrent problem in
applications that rely on correct results for their operation (e.g. \cite{jose-pldi11,safarpour-fmcad07}).
If a SAT solver returns an assignment and reports it to be
{\em SATISFIABLE}, then to check if the assignment is indeed a satisfying
assignment, it is enough to go through each clause, and check
if the clause is satisfied by the assignment.
% (that is, check if there is a literal in the clause assigned \emph{true}).
%
If all the clauses are satisfied, then the assignment is a satisfying
assignment.
Otherwise, the result is incorrect.

%If the SAT solver reports the instance to be {\em UNSATISFIABLE}, then the
%case is not as simple to validate.
%
Similarly, it is necessary to validate {\em UNSATISFIABLE} outcomes.
In 2003, Zhang \& Malik~\cite{zhang-date03} proposed an independent
resolution based checker that takes the trace produced by the solver
and checks the correctness of the result.
Goldberg et al.~\cite{goldberg-date03} considers a procedure for the
verification of unsatisfiable formulas
and in 2009 Weber et al.~\cite{weber-jal09} proposed the use a HOL theorem
prover to verify the proofs of unsatisfiability given by minisat~\cite{minisat}
and zchaff~\cite{chaff}.

In the context of Boolean optimization,
%, the result provided by a solver is not a satisfiable or unsatisfiable result.
%
the solver needs to provide the optimum value it has found and
certificates that the value is correct.
Recently, Larrosa et al.~\cite{larrosa-jar09,larrosa-sat09}
showed how to generate \emph{proofs of optimality} for
branch-and-bound procedures that corresponds to a lower-bound
certificate and a model to the optimization problem.

This section shows how to modify the algorithms described in the
previous section, so that their results can be validated.
The objective is to instrument Boolean optimizers to return a uniform
certificate with the minimal information, that allows an independent
checker to validate the result. 
Two methods of validation are proposed.

\subsection{Method 1}

All the algorithms of the previous section are based in iteratively
searching through CNF instances, either satisfiable or unsatisfiable.
The value being optimized is encoded in the CNF instances.
The first method to validate the result returned by these algorithms is
to consider the value encoded and the results returned by the
SAT solver for all iterations, that is, to validate the result returned
by the SAT solver for each iteration.

For validation of iterations with satisfiable CNF instances, the
solvers report the value that is being
tested and the satisfiable assignment returned by the SAT solver.
For example, consider the Linear search algorithm going through
unsatisfiable instances of the previous
section and an instance $\varphi$.
Suppose that the algorithm runs for three iterations (all with
unsatisfiable results) and that on the fourth iteration the SAT solver
reports the instance to be satisfiable.
The algorithm would report for the fourth iteration the bound $\lambda=3$
meaning that the
total number of satisfied clauses is 
$|\varphi|-3$, together with the satisfying assignment.
In this case, the check needs only to verify that there are exactly
three clauses that are not satisfied by the assignment reported.

For the case of unsatisfiable CNF instances, the Boolean optimizers
need only to report the trace produced by the SAT solver (as in the
case of checking unsatisfiability of SAT
solvers~\cite{zhang-date03}).

An independent checker receives the information of the satisfiable
iterations, which we call the \emph{satisfiable certificates}, and the traces of
the unsatisfiable iterations (the \emph{unsatisfiable certificates}), and
validates the result.
The checker verifies that the satisfiable certificates are correct, that is,
the assignment is a satisfiable assignment and satisfies
the value reported.
For the unsatisfiable certificates the checker proceeds as current resolution
checkers of SAT solvers in unsatisfiable instances as in Zhang \& Malik~\cite{zhang-date03}.

\begin{example}
\label{ex:method1}

  Consider the following partial MaxSAT formula:
  \begin{displaymath}
    \begin{array}{ll}
      \textrm{Soft CLauses:} & (\neg x_1) (\neg x_2) (\neg x_3)\\
      & (\neg x_4) (\neg x_5) \\
      \textrm{Hard Clauses:} & (x_1\vee x_2) (x_2\vee x_3) \\
      & (x_3\vee x_4)(x_4\vee x_5)\\
      & (x_1\vee x_5)
    \end{array}
  \end{displaymath}
  
  Consider for the example a Linear search algorithm.
  A correct Linear search algorithm going through unsatisfiable instances
  would start by relaxing the five soft clauses and then perform
  four iterations. 
  The first three iterations (with $\lambda=0,1,2$) would each report an
  unsatisfiable certificate, while the last iteration (with
  $\lambda=3$) would report a satisfiable certificate.
 
  The checker using the previous method 1 would have to validate the four
  certificates. 
\end{example}

\subsection{Method 2}

Given that the iterations of the approaches described for Boolean optimization
converge to the optimum value through satisfiable and unsatisfiable
instances, then in the second  method of validation, not all the
certificates of all the iterations are checked. 
In fact, the checker needs only to validate the {\em last} satisfiable
iteration and the {\em last} unsatisfiable trace produced.
This can easily be seen for the Linear Search algorithms of
Section~\ref{sec:alg-linear-search}.
Consider a run of the Linear search algorithm moving through
unsatisfiable instances.
In terms of the variable $\lambda$ and the status returned by
the SAT solver $st$, the run of the Linear search algorithm looks like
the following: 
\begin{displaymath}
  \begin{array}{ll}
    \lambda = 0 & st: \mathit{UNSATISFIABLE};\\
    \ldots\\
    \lambda = k-1 & st: \mathit{UNSATISFIABLE};\\
    \lambda = k & st: \mathit{SATISFIABLE}
  \end{array}
\end{displaymath}

The last unsatisfiable iteration has $\lambda = k-1$.
If we check that the SAT solver returned the correct unsatisfiable result for
the formula $\varphi_W~ \cup~ CNF(\sum_{r\in R}r\leq k-1)$ of the
last unsatisfiable iteration, then we are
assured that any of the previous iterations with $\lambda < k-1$ are
all unsatisfiable.
This is true because the formula $\varphi_W$ and the set of relaxation
variables remains the same between iterations, and thus the set of
solutions of the constraint $(\sum_{r\in R}r\leq \lambda)$, with
$\lambda < k-1$, is included in the set of solutions of the constraint
$(\sum_{r\in R}r\leq k-1)$.  
Only one satisfiable iteration exists with
$\lambda = k$, which corresponds to the optimal value.

In the case of the Linear search algorithm going through satisfiable
instances, it is also easy to demonstrate that it is enough to check the
certificates of to the last satisfiable and the last unsatisfiable
iterations. 
Consider a run of the Linear search algorithm going through satisfiable
instances.
As before, we consider the iterations in terms of the variable $\mu$ and the status returned by
the SAT solver $st$.
Consider w.l.o.g. the worst case scenario where the assignment returned
$\sigma$ always decreases $\mu$ by one unit.
Then a run of the Linear search algorithm (through
satisfiable instances) looks like the following:
\begin{displaymath}
\begin{array}{ll}
  \mu = n  & st: \mathit{SATISFIABLE};\\
  \ldots\\
  \mu = n-k+1 &  st: \mathit{SATISFIABLE};\\
  \mu = n-k & st: \mathit{UNSATISFIABLE}\\
\end{array}
\end{displaymath}

There is only one unsatisfiable iteration with $\mu =
n-k$ and it corresponds to the last unsatisfiable certificate.

The last satisfiable iteration has $\mu = n-k+1$ (the optimum value).
If we check that the SAT solver returned the correct satisfiable
result for the formula $\varphi_W~ \cup~ CNF(\sum_{r\in R}r\leq
n-k+1)$ of the last satisfiable iteration, then we are assured that
for greater values of $\mu$, the formula is still satisfiable.
The reasons for this are analogous to the previous case of Linear
search going through unsatisfiable instances.

The case of the Binary search algorithm of Section~\ref{sec:alg-bin-search} 
is similar to the Linear search algorithms but using $\mu$
and $\lambda$.
The algorithm terminates with $\mu =
\lambda+1$, and as in the Linear search algorithms the last
unsatisfiable and the last satisfiable iterations will subsume the
other iterations, for analogous reasons.

\begin{example}
  Consider one  more time the instance of the previous Example~\ref{ex:method1}.

  A checker using method 2 for validating would not have to validate
  all the unsatisfiable certificates.
  Instead, using method 2 would save the checker from validating the
  first two unsatisfiable certificates.
  Only the last unsatisfiable certificate and the satisfiable
  certificate would have to be validated.
\end{example}

An interesting case for validating is the case of the
simplified MSU3 algorithm, which changes its set of relaxation
variables while it is changing the bound $\lambda$.
Due to the change of the set of relaxation variables it is not
possible to consider only the last \emph{satisfiable certificate} and the last
\emph{unsatisfiable certificate} as in the other algorithms.

For example, consider the MaxSAT instance $\{(x)\wedge()\}$ which has
an optimum value of $\lambda = 1$.
Consider also a buggy simplified MSU3 solver with the following
run:
\begin{displaymath}
\begin{array}{ll}
  \lambda = 0~   st: \mathit{UNSATISFIABLE}& core: \{(x)\};\\
  \lambda = 1~  st: \mathit{UNSATISFIABLE}& core: \{()\};\\
  \lambda = 2~ st: \mathit{SATISFIABLE} & \sigma=\{x=r_1=r_2=1\}\\
\end{array}
\end{displaymath}

On the first iteration, the solver correctly returns
{\em UNSATISFIABLE} but with the wrong core $\{(x)\}$.
Clause $(x)$ is augmented with the relaxation variable $r_1$ being
transformed into $(x \vee r_1)$.
Thus on the second iteration the solver tests the satisfiability of the
CNF instance $\{(x\vee r_1)\wedge()\wedge CNF(r_1\leq 1)\}$, and
returns {\em UNSATISFIABLE} with the core $\{()\}$. 
Clause $()$ is relaxed with relaxation variable $r_2$ which becomes
$(r_2)$.
Finally on the third iteration the solver tests the satisfiability of
the instance $\{(x\vee r_1)\wedge(r_2)\wedge CNF(r_1+r_2\leq 2)\}$,
and reports {\em SATISFIABLE} with an assignment $\sigma=\{x=r_1=r_2=1\}$ and $\lambda=2$.

The last unsatisfiable certificate, and the last satisfiable
certificate are both correct and yet the result is wrong.
This example shows that when modifying the set of relaxation
variables, it is not enough to check the last two certificates.
An additional test is required, to test that among all possible (soft)
clauses to relax, the value reported is minimal.
This can be achieved by creating a new instance with all (soft)
clauses relaxed together with a new constraint that encodes the sum of all
relaxation variables being strictly smaller than the result returned by the
Boolean optimization solver.

In the previous example the checker would validate both the last satisfiable
and unsatisfiable certificates and make the test that the CNF
instance $\{(x \vee r_1) \wedge (r_2) \wedge CNF (r_1+r_2<2)\}$ is
unsatisfiable. 
The SAT solver would report the instance to be satisfiable, with the
satisfying assignment $\{x=r_2=1,r_1=0\}$, and the checker would
report the MaxSAT solver to have an incorrect result.

Notice that we could have restricted the original satisfying assignment
$\sigma$ (reported on the satisfiable certificate) to an assignment $\sigma'$ containing only assignments to the
original variables ($\sigma'=\{x=1\}$) and then counted the
number of clauses which are not satisfied by original variables.
The obtained value could then be compared with the reported
$\lambda$.
For the previous example we would have obtained that only one clause
is unsatisfied, thus realizing that the reported $\lambda=2$ is not
minimal.
But this is not always the case, if the reported $\sigma$ was instead
$\sigma=\{x=0,r_1=r_2=1\}$, then the number of clauses not satisfied
by original variables would also be two as $\lambda$, thus still
requiring the additional test to verify that the reported $\lambda$ is minimal.

The correctness of the second method of validation can be summarized
in the following propositions.

\begin{proposition}[Validation of Linear/Binary Search]

The result of a Boolean Optimizer based on Linear search or on Binary
search is correct if and only if the last unsatisfiable certificate and the
last satisfiable certificate are validated.

\end{proposition}

%\begin{theorem}
%The result of the simplified MSU3 algorithm can be shown correct by
%considering a modified SAT test (described above).
%%
%The result of the modified MSU3 algorithm is
%correct if and only if the last unsatisfiable certificate (of the modified
%problem instance) and the last satisfiable certificate are validated
%together with. 
%%
%\end{theorem}

\begin{proposition}[Validation of the Simplified MSU3]

The result of the simplified MSU3 algorithm is
correct if and only if the last unsatisfiable certificate (of the modified
problem instance) and the last satisfiable certificate are validated
together with the validation that the value returned is minimal.

\end{proposition}

%In the previous therorem, the validation of the returned value being
%minimal can be done using an aditional call to a SAT solver.

%------------------------------------------------------------------------------%

%------------------------------------------------------------------------------%
% File:        results.tex
%
% Description: Experimental results.
%
% Author:      Joao Marques-Silva.
%
% Created:     07 May 2009.
%------------------------------------------------------------------------------%

\section{Experimental Results}
\label{sec:res}

\begin{table*}[t]
  \begin{center}
    \small
    \begin{tabular}{@{}|@{}c@{}|@{}c@{}|@{}c@{}|@{}c@{}|@{}c@{}|@{}c@{}|@{}}
      \hline
      {\scriptsize Instance} & {\scriptsize Opt. Value} & {\scriptsize Bin Search} &
          {\scriptsize Bin Search-GC} & {\scriptsize Check All} & {\scriptsize Check One}\\
      \hline\hline
      {\scriptsize simp-unif-100\_100.09.wcnf} & 26 & 1.58 & 1.67  & 0.17 & 0.01\\
      \hline
      {\scriptsize simp-ibd\_50.03.wcnf} & 54 & 6.04 & 6.62  & 1.34 & 0.01\\
      \hline
      {\scriptsize normalized-aim-100-1\_6-yes1-1.wcnf} & 100 & 2.07 & 2.12  & 0.02 & 0.02\\
      \hline
      {\scriptsize normalized-aim-200-1\_6-yes1-2.wcnf} & 200 & 11.28 & 11.34 & 0.04 & 0.04 \\
      \hline
      {\scriptsize normalized-ii8a1.wcnf} & 54 & 37.07 & 43.15 & 23.42 & 0.07 \\
      \hline
      {\scriptsize normalized-jnh1.wcnf} & 92 & 8.39 & 9.25 & 2.05 & 0.07 \\
      \hline
      {\scriptsize normalized-jnh213.wcnf} & 92 & 3.02 & 3.11 & 0.31 & 0.02 \\
      \hline
      {\scriptsize normalized-jnh7.wcnf} & 89  & 3.95 & 4.23 & 0.76 & 0.01 \\
      \hline
      {\scriptsize normalized-par8-1.wcnf} & 350 & 41.48 & 41.47  & 0.22 & 0.22\\
      \hline
    \end{tabular}
  \end{center}
  \caption{Run and checking times for the Binary Search algorithm}
  \label{tab:solvers-results}
\end{table*}

This section presents experimental results on checking MaxSAT
solutions computed with Binary Search algorithm. 
The methods outlined in the paper could be
used with the other Boolean optimization approaches studied in this
paper. 
%However, as described earlier, core-based MaxSAT poses the most
%interesting challenges.
%
Two types of results are presented.
The first type of results show the CPU times of the solver with and without
the generation of the certificates.  
The second type of results concentrate on checking all the certificates
versus checking just one certificate.
All the results obtained are for the Binary Search 
algorithm described in Section~\ref{sec:alg-bin-search}. 
The nine instances used in the results were obtained from the 2009
MaxSAT evaluation, and represent partial MaxSAT instances.
A Mac Pro server with 8GByte of physical memory and a 2.93GHz
processor was used for the experiments. All run times are in seconds.

The values in the columns Bin Search and Bin Search-GC of
Table~\ref{tab:solvers-results} show the running times of the two 
versions of the solver, with and without the generation of
certificates.
The table also shows the optimum value for each problem instance.
As can be concluded, there can be a difference in run times between
generating and not generating certificates. Similar conclusions were
observed in~\cite{larrosa-jar09,larrosa-sat09}.

The values in the columns Check ALL and Check One of
Table~\ref{tab:solvers-results} represent the running times of
checking  all the unsatisfiability certificates, and of checking just
the last unsatisfiable certificate\footnote{The
  running times of checking the satisfiable certificate is negligible and was
  not considered in either approaches.}.
From the results we can conclude that as expected checking only the
last unsatisfiable certificate can result in significant savings in
terms of run times.
For example, the normalized-iia81.wcnf becomes two orders of
magnitude faster than considering all the certificates. Similar
conclusions can be drawn for most of the other benchmarks shown.
Nevertheless, some instances show the same time on checking all or
just one. This happens on instances with just one unsatisfiable
iteration, thus just one unsatisfiable certificate to check.
The results allow concluding that the ability to check just one
unsatisfiable certificate and one satisfiable certificate can result
in important performance improvements when checking the results of
Boolean optimization solvers.

%------------------------------------------------------------------------------%

%------------------------------------------------------------------------------%
% File:        conc.tex
%
% Description: Conclusions.
%
% Author:      Joao Marques-Silva.
%
% Created:     07 May 2009.
%------------------------------------------------------------------------------%

\section{Conclusions and Future Work}
\label{sec:conc}

This paper investigates solutions for checking the results
%solutions
computed
by Boolean optimizers, which are based on iterative calls to a SAT
solver. Hence, the paper complements recent work on generating
certificates for branch-and-bound Boolean optimization algorithms.
The paper overviews all existing algorithms based on iterative calls
to a SAT solver, and shows that, for all these algorithms, it suffices
to check one unsatisfiability proof and one satisfiable certificate.
Experimental results indicate that the overhead of checking the
solutions computed by Boolean Optimization algorithms is negligible.
Simple implementation improvements to the work described in the paper
include eliminating altogether proof tracing, only recreating proof
tracing for the (then known) last unsatisfiability proof. This
provides additional performance improvements over solutions that might
check all unsatisfiability proofs.
Additional research work consists in developing solution checking
approaches for Max-SMT~\cite{nieuwenhuis-sat06} and
Max-ASP~\cite{emilia-lpnmr09}.

\begin{comment}
ajrm::For future work:
\begin{itemize}
  \item Expand the checking for Weighted Max-SAT.
    Need to consider the weights of each clause.

  \item Expand the checking for Max-SMT.

  \item Expand the checking for Max-ASP.
\end{itemize}
\end{comment}

%------------------------------------------------------------------------------%

\section*{Acknowledgments}

This work is partially supported by SFI PI grant BEACON (09/IN.1/I2618) and by
FCT through grant ATTEST (CMU-PT/ELE/0009/2009).

\bibliographystyle{splncs03}
\bibliography{xrefs}

\end{document}